\documentclass[conference]{IEEEtran}
\IEEEoverridecommandlockouts
\usepackage{cite}
\usepackage{amsmath,amssymb,amsfonts}
\usepackage{algorithmic}
\usepackage{graphicx}
\usepackage{textcomp}
\usepackage{xcolor}
\def\BibTeX{{\rm B\kern-.05em{\sc i\kern-.025em b}\kern-.08em
    T\kern-.1667em\lower.7ex\hbox{E}\kern-.125emX}}
\begin{document}

\title{PAT: Parallel Attention Transformer for Visual Question Answering in Vietnamese}

% \author{\IEEEauthorblockN{1\textsuperscript{st} Nghia Hieu Nguyen}
% \IEEEauthorblockA{\textit{Department of Information Science and Engineering, University of Information Technology, Ho Chi Minh City, Vietnam} \\
% \textit{Vietnam National University, Ho Chi Minh City, Vietnam} \\
% 19520178@gm.uit.edu.vn}
% \and
% \IEEEauthorblockN{2\textsuperscript{nd} Kiet Van Nguyen}
% \IEEEauthorblockA{\textit{Department of Information Science and Engineering, University of Information Technology, Ho Chi Minh City, Vietnam} \\
% \textit{Vietnam National University, Ho Chi Minh City, Vietnam}\\
% kietnv@uit.edu.vn}
% }

\author{Nghia Hieu Nguyen$^{1,2,3}$, Kiet Van Nguyen$^{1,2,4}$ \\
$^{1}$Faculty of Information Science and Engineering, University of Information Technology, Ho Chi Minh City, Vietnam \\
$^{2}$Vietnam National University, Ho Chi Minh City, Vietnam \\
Email: $^{3}$19520178@gm.uit.edu.vn, $^{4}$kietnv@uit.edu.vn}

\maketitle

\begin{abstract}
We present in this paper a novel scheme for multimodal learning named the Parallel Attention mechanism. In addition, to take into account the advantages of grammar and context in Vietnamese, we propose the Hierarchical Linguistic Features Extractor instead of using an LSTM network to extract linguistic features. Based on these two novel modules, we introduce the Parallel Attention Transformer (PAT), achieving the best accuracy compared to all baselines on the benchmark ViVQA dataset and other SOTA methods including SAAA and MCAN.
\end{abstract}


\section{Introduction}
Multimodal learning recently attracted lots of attention from the research community because of its exciting and challenging requirements. Multimodal learning aims to explore how to extract and fuse multimodal information effectively. Typical tasks of multimodal learning can be listed as Visual Question Answering (VQA) where a machine is required to answer a given question based on visual information from a given image \cite{VQA}, Image Captioning (IC) where  a machine is required to generate natural language captions that describe the content of the given image \cite{VQA}, or Visual Grounding where a machine is required to draw bounding boxes on images that indicate objects mentioned in a given query using natural language \cite{yu2016modeling}. 

Most attention concentrates on the multimodal tasks relevant to visual-textual information, particularly the VQA task. Current approaches on VQA treat this task as an answers classification task. This guide the development of VQA methods focusing on studying the most effective scheme to fuse information from the given image and question in order to select the best accurate candidate among a given set of answers. According to the survey study of Zhang et al. \cite{ZHANG2019268}, based on the way of performing attention, VQA methods can be grouped into two types: single-hop attention methods and multi-hop attention methods. On large benchmark VQA for English, various works show that single-hop attention methods do not achieve good results compared to multi-hop attention methods. 

In this paper, we present a new multi-hop attention method for fusing information from images. Our experimental results prove that single-hop attention methods find difficulty when they tackle the VQA even on a small-size dataset as ViVQA \cite{tran-etal-2021-vivqa-vietnamese}. 
\section{Related works}
\subsection{VQA datasets}
Antol et al. \cite{VQA} first introduced the VQA task by releasing the VQAv1 dataset. This dataset includes 254,721 images with 764,163 questions and 4,598,610 answers. Most of the attention is drawn to the VQAv1 dataset \cite{kazemi2017show,yu2019deep,goyal2017making,teney2018tips} and many attention mechanisms were proposed that still affect the mindset of design for later methods \cite{lu2016hierarchical,kazemi2017show,yu2019deep} such as Co-Attention \cite{lu2016hierarchical} and Stacked Attention \cite{kazemi2017show}.

Results of former studies on the VQAv1 dataset achieved pretty good results \cite{teney2018tips} by treating the VQA task as answer selection over a defined set of candidates or answer classification. However, as other classification tasks, answer imbalance in the VQAv1 dataset forms a novel problem that was indicated by Goyal et al. \cite{goyal2017making}. Goyal et al. \cite{goyal2017making} proved that former VQA methods obtained good results on the VQAv1 dataset as they suffered from the language prior methods. Particularly, when being given a question, former VQA methods recognize its pattern and select the most apparent answer belonging to that pattern as the candidate, despite the visual information of the images.

To overcome the language prior phenomenon, Goyal et al. \cite{goyal2017making} balanced the VQAv1 datasets and then proposed the VQAv2 dataset. Goyal et al. \cite{goyal2017making} constructed lots of experiments and showed that former VQA methods did not perform well as they had behaved. The VQAv2 dataset contains 204,721 images with 1,105,904 questions and 11,059,040 answers, which becomes the largest benchmark for the VQA task in English.

Recent studies constructed VQA datasets that required reading comprehension of VQA methods \cite{singh2019towards,mishraICDAR19,Mathew_2021_WACV}. Moreover, to develop a VQA system that can use incorporate knowledge while answering the given questions, lots of datasets were released \cite{marino2019ok}. On the other side, former VQA methods were designed to select answers rather than forming sentences to answer as humans. From that on, there are works conducted the open-ended VQA datasets \cite{Kantharaj2022OpenCQAOQ,Tanaka_Nishida_Yoshida_2021} to research the answer-generation methods instead of answer-selection ones.

In Vietnamese, the first VQA dataset was introduced by Tran et al. \cite{tran-etal-2021-vivqa-vietnamese}. This dataset was constructed based on the COCO-QA dataset \cite{lin2014microsoft} using a semi-automatic method. Recently, Nguyen et al. \cite{nguyen2023vlsp} introduced the multilingual VQA dataset, the UIT-EVJVQA dataset, in three languages Vietnamese, English, and Japanese. This dataset is the first open-ended VQA dataset that includes Vietnamese. In addition, Nghia et al. \cite{nguyen2023openvivqa} presented a Vietnamese open-ended VQA dataset consisting of 11,000+ images associated with 37,000+ question-answer pairs (QAs).

\subsection{VQA methods}
Former VQA methods were designed based on the attention mechanism \cite{vaswani2017attention}. One well-known baseline on the VQAv1 dataset is the Hierarchical Co-Attention Network \cite{lu2016hierarchical} which used the Convolutional Neural Network (CNN) \cite{cnn} to extract the n-gram features from questions and used the co-attention to perform attention mechanism over questions and images. Later studies based on this co-attention proposed various methods such as ViLBERT \cite{lu2019vilbert}, VisualBERT \cite{li2019visualbert}, or LXMERT \cite{tan2019lxmert}.

Another strong baseline on the VQAv1 dataset proposed by Kazemi et al. \cite{kazemi2017show} introduces the Stack Attention. This kind of attention stacks the visual features and linguistic features together and then yielded the attention map over the two kinds of features. Later work proposed methods based on Stack Attention but using transformer \cite{vaswani2017attention} such as VL-BERT \cite{su2019vl},
Unicoder-VL \cite{li2020unicoder}, Uniter \cite{chen2020uniter}, X-LXMERT \cite{cho2020x}, Pixel-BERT \cite{huang2020pixel}, or VLMo \cite{bao2021vlmo}.
\section{Our proposed method}

\begin{figure*}
    \centering
    \includegraphics[width=0.5\textwidth]{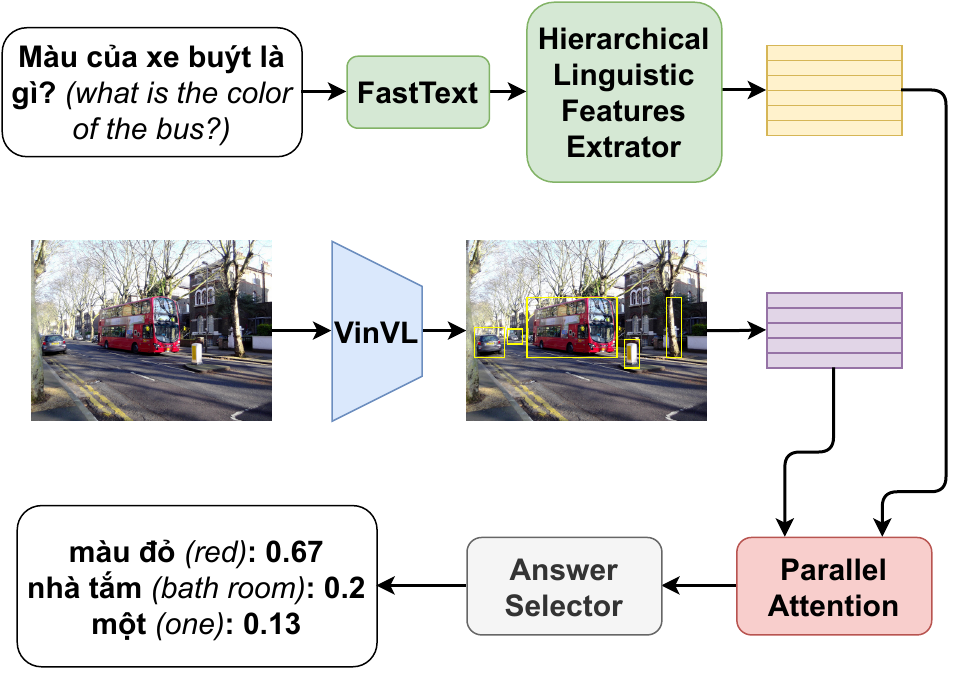}
    \caption{Overall structure of the PAT method.}
    \label{fig:overall}
\end{figure*}

Inspired by the success of the transformer \cite{vaswani2017attention} and the study of Yu et al. \cite{yu2019deep}, we propose a novel scheme of attention, Parallel Attention, that is a kind of multi-hop attention and differs from recent methods. Moreover, to leverage the linguistic features of Vietnamese, we provide Parallel Attention with the hierarchical feature extractor for questions and hence propose a novel method, the Parallel Attention Transformer (PAT). Our experiments prove that this hierarchical extractor is indeed necessary.

The PAT method includes four main components: the Hierarchical Linguistic Feature Extractor, the Image Embedding module, the Parallel Attention module, and the Answer Selector (Figure \ref{fig:overall}). The detailed architecture of our method is detailed as follows.

\subsection{Hierarchical Linguistic Feature Extractor} \label{sect:hierarchical}

\begin{figure}
    \centering
    \includegraphics[width=0.5\textwidth]{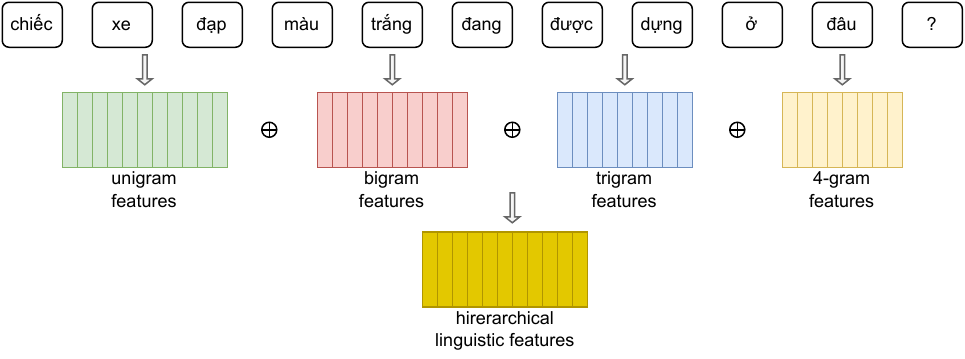}
    \caption{Hierarchical Linguistic Features Extractor.}
    \label{fig:hierarchical}
\end{figure}

We apply a pre-trained word embedding for Vietnamese to extract the linguistic features of questions. As each token of questions after being passed through the pre-trained word embeddings they are mapped to respective embedded vectors. Accordingly extracted features using word embedding are the unigram features. We aim to make our method have the ability to fully catch the linguistic context of the sentence, so we propose to construct the n-gram linguistic features based on the unigram features (Figure \ref{fig:hierarchical}).

Particularly, we use a 1D convolutional neural network (CNN) with a kernel of size 1, 2, 3, and 4 to extract the unigram, bigram, trigram, and 4-gram of the initial unigram features, respectively. We note that as the initial unigram features of pre-trained word embedding might not be in the same latent space of the model, so we use a 1D CNN with the kernel of size 1 to project these features into latent space. Our ablation study will prove that this 1D CNN is important to improve the accuracy of our proposed method. The four n-gram features finally are summed to yield the hierarchical linguistic features for questions.

\subsection{Image Embedding module}

Inspired by the study of Anderson et al. \cite{Anderson2017BottomUpAT}, we perform the bottom-attention mechanism on the visual features. Particularly, we used the VinVL pre-trained image models \cite{zhang2021vinvl} to achieve the region features from images. The VinVL pre-trained model was trained on large-scale datasets of vision-language tasks and they used detected tags of objects together with the ROI features of Faster-RCNN-based models hence their visual features are rich in visual aspect as well as linguistic aspect, and Zhang et al. \cite{zhang2021vinvl} proved that VinVL outperformed previous pre-trained image models on various tasks. The visual features are projected into the latent space of the model before being passed to the next components by using a fully connected layer.

\subsection{Parallel Attention module.}

\begin{figure}
    \centering
    \includegraphics{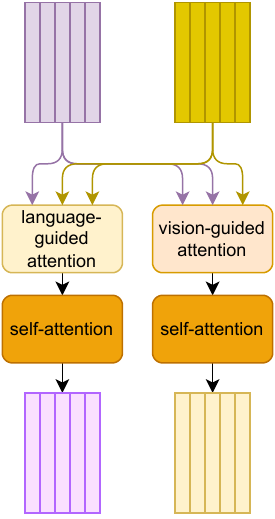}
    \caption{A Parallel Attention module.}
    \label{fig:pat}
\end{figure}

As various VQA models \cite{yu2019deep,li2019visualbert,kim2021vilt,lu2019vilbert,lu2022unified,chen2020uniter,li2020unicoder}, our proposed method has an encoder module containing encoder layers to perform the attention mechanisms. In particular, the Parallel Attention module has four components. Each component has an attention layer \cite{vaswani2017attention} and a Position-wise Feed Forward layer \cite{vaswani2017attention}.

The attention layer is the multi-head attention module proposed by Vaswani et al. \cite{vaswani2017attention}. Given a query $Q$, key $K$, and value $V$ vector, the attention map is specified as follows:
\begin{equation}
    A = softmax(\frac{QK^T}{\sqrt{d_k}})
\end{equation}
where $d_k$ is the dimension of the value vector and we assume that $Q$, $K$, and $V$ have the same dimension. After obtaining the attention map, the encoded features are determined as:
\begin{equation}
    Y = A V
\end{equation}

In the Parallel Attention module, the first two components are used to perform cross-and-parallel attention: vision over language and language over vision, respectively, by changing the query, key, and value role of visual features and linguistic features. The last second components are used to perform self-and-parallel attention: vision over itself and language over itself by defining the key, query, and value are all visual features or linguistic features (Figure \ref{fig:pat}). Finally, the visual features $x_v$ and linguistic features $x_l$ are produced that have advantage information for selecting an accurate candidate among defined answers.

\subsection{Answer Selector}
The Answer Selector module is designed to fuse the information of visual features $x_v$ and linguistic features $x_l$ that produce the fused features $x_f$. The fused features are then projected into the vocab space. Finally, we obtained the probabilistic vector that indicates the most candidate as an answer. We follow the Attribute Reduction and Classifier of MCAN \cite{yu2019deep} method to design the Answer Selector.

In particular, the Answer Selector module includes two phases: attributes reduction and Candidate Selection (in the context of the study of Yu et al. \cite{yu2019deep}, this phase is named Answer Classifier). Given $x_v$ and $x_f$ obtained from the Parallel Attention layers, we use the MLP layer with the softmax function to re-weight these features:
\begin{equation}
    attr_v = softmax(MLP(x_v))
\end{equation}

\begin{equation}
    attr_l = softmax(MLP(x_l))
\end{equation}

Then the reduced attributes are applied to denoise and combine the visual features $x_v$ and linguistic features $x_l$:

\begin{equation}
    x_v = sum(x_v * attr_v)
\end{equation}

\begin{equation}
    x_l = sum(x_l * attr_l)
\end{equation}
where $*$ indicates the element-wise product.

Finally, the fused features $x_f$ are obtained by summing the $x_v$ and $x_l$:
\begin{equation}
    x_f = W_v x_v + W_l x_l
\end{equation}

The selected candidate $c$ is determined based on the fused features $x_f$:
\begin{equation}
    c = max(W_{vocab} x_f)
\end{equation}
\section{Experimental results}

\subsection{Dataset}
In this paper, we propose the Hierarchical Linguistic Feature Extractor to leverage the advantages of grammar and context in Vietnamese. Accordingly, we conduct experiments on the ViVQA dataset \cite{tran-etal-2021-vivqa-vietnamese} which is the first visual question answering dataset for Vietnamese.

\subsection{Evaluation Metrics}
We follow the study of Teney et al. \cite{teney2018tips} that treats the VQA task as a classification task. From that on, we use the Accuracy metric or Exact Match (EM) metric defined by Antol et al. \cite{VQA} to measure the ability of VQA methods in our experiments. Particularly, the EM metric is determined as:

\begin{equation}
    EM = \frac{1}{n} \sum_{i=1}^{n} \left( \frac{1}{m} \sum_{j = 1}^{m} \left(1 -\alpha_{ij} \right) \right)
\end{equation}

\begin{equation}
    \alpha_{ij} = 
    \begin{cases}
        0 \Leftrightarrow \hat{a_i} = a_{ij} \\
        1 \text{ otherwise}
    \end{cases}
\end{equation}
where $n$ is the total number of questions in whole dataset, $m$ is the total number of answers of given question $i$, $\hat{a_i}$ is the predicted answer for question $i$, $a_{ij}$ is the $j$th ground truth answer for question $i$.

\subsection{Baselines}
We compare our proposed PAT method with all models implemented in the previous work \cite{tran-etal-2021-vivqa-vietnamese}. In addition, we re-implemented the two baselines on VQAv1 and VQAv2 datasets, which are SAAA \cite{kazemi2017show} and MCAN \cite{yu2019deep} methods, respectively.

\subsection{Configuration}
All experiments in this paper used the VinVL pre-trained image \cite{zhang2021vinvl} to extract region features \cite{Anderson2017BottomUpAT} and grid features \cite{jiang2020defense}. Both SAAA and MCAN as well as PAT use FastText \cite{bojanowski2017enriching} as pre-trained word embeddings to extract features of questions. All implemented experiments were performed on an A100 GPU, with batch size 64 and the learning rate fixed at 0.01. We used Adam \cite{Kingma2014AdamAM} as the optimization method. The detailed configuration for each method is listed as follows:

\subsubsection{SAAA (Show, Asked, Attend, and Answer)}
We followed the configuration of SAAA that made this model obtain the best results on VQAv1 \cite{kazemi2017show}. In particular, the LSTM \cite{lstm} layer of SAAA has 1024 as its hidden dimension, and the attention size is 512. In the Classifier module of SAAA, features are mapped into 1024-dimensional space before being projected into the vocab space. In our implementation, we used VinVL instead of ResNet152 \cite{he2016deep} to achieve the grid features.

\subsubsection{MCAN (Deep Modular Co-Attention Network)}
We followed the best configuration of MCAN reported in the study \cite{yu2019deep}. In particular, we used 6 layers for the Co-Attention module. The multi-head attention modules of MCAN have 512 as their hidden size. We used VinVL to extract region features instead of Faster-RCNN \cite{ren2015faster}.

\subsubsection{PAT}
The Hierarchical Linguistic Feature Extractor contains 4 CNN layers with respectively 1, 2, 3, and 4 as their kernel size to extract unigram, bigram, trigram, and 4-gram features. The Parallel Attention module contains 4 layers. All attention modules of each layer in the Parallel Attention module have 512 as their hidden dimension. We follow \cite{Devlin2019BERTPO} to use GeLU as an activation function instead of ReLU as in \cite{vaswani2017attention}.

\subsection{Results}

\begin{table}[ht]
    \centering
    \caption{Experimental results on the ViVQA dataset. Note that (*) indicates results from \cite{tran-etal-2021-vivqa-vietnamese}.}
    \begin{tabular}{lr}
        \hline
        \textbf{Methods}                & \textbf{EM}                       \\ \hline
        LSTM + W2V (*)                      & 0.3228                              \\
        LSTM + FastText (*)                 & 0.3299                              \\
        LSTM + ELMO (*)                    & 0.3154                              \\
        LTSM + PhoW2Vec (*)                & 0.3385                              \\
        Bi-LSTM + W2V (*)                  & 0.3125                              \\
        Bi-LSTM + FastText (*)             & 0.3348                              \\
        Bi-LSTM + ELMO (*)                 & 0.3203                              \\
        Bi-LTSM + PhoW2Vec (*)             & 0.3397                              \\
        Hierarchical Co-Attention + LSTM (*) & 0.3496                              \\ \hline
        SAAA                            & 0.5415                              \\
        MCAN                            & \multicolumn{1}{l}{0.5711}          \\
        \textbf{PAT (ours)}                  & \multicolumn{1}{l}{\textbf{0.6055}} \\ \hline
    \end{tabular}
    \label{tab:result}
\end{table}

As indicated in Table \ref{tab:result}, SAAA and MCAN achieved significantly better results compared to all implementations of Tran et al. \cite{tran-etal-2021-vivqa-vietnamese}. Straightforward structures such as the combination of pre-trained word embeddings and LSTM \cite{lstm} do not tackle effectively such complicated tasks as VQA, while deeper and ingeniously designed methods such as SAAA and MCAN took over the ViVQA dataset better.

Especially, our proposed method, PAT, obtained the best results while leaving other methods a far distance. Particularly, PAT achieved approximately 6\% better than SAAA and approximately 3\% better than MCAN despite these two methods are the SOTA methods on the VQAv1 and VQAv2 that were not pre-trained on large-scale datasets.

\subsection{Ablation study}

\begin{table}[ht]
    \centering
    \caption{Ablation study for PAT method.}

    \begin{tabular}{lr}

        \hline
        \textbf{Methods}    & \textbf{EM}   \\ \hline
        PAT w/o Hier.     & 0.5868          \\
        PAT w LSTM &      0.5981     \\
        \textbf{PAT}      & \textbf{0.6055} \\ \hline
    \end{tabular}
    \label{tab:ablation}
\end{table}

we conduct an ablation study to comprehensively discover how our two proposed modules, Hierarchical Linguistic Feature Extractor, and Parallel Attention module, contribute to the overall result of the PAT. Results are shown in Table \ref{tab:ablation}.

According to Table \ref{tab:ablation}, the PAT which does not use LSTM or Hierarchical Linguistic Features Extractor to extract features of questions obtained lower accuracy. When equipped with LSTM or Hierarchical Linguistic Extractor, PAT achieved better results. Especially it achieved the best results when using the Hierarchical Linguistic Extractor. This result proves that the Hierarchical Linguistic Feature Extractor leverages the grammar dependency as well as the context of Vietnamese better than a simple LSTM network.

\begin{table}[ht]
    \centering
    \caption{Ablation study for PAT that use 1-size kernel CNN to extract unigram features}
    \label{tab:ablation_cnn}
    \begin{tabular}{lr}
        \hline
        \textbf{Methods}   & \textbf{EM} \\ \hline
        PAT w/o 1-size kernel CNN & 0.5848        \\
        PAT w 1-size kernel CNN   & 0.6055        \\ \hline
    \end{tabular}
\end{table}

Moreover, as stated in Section \ref{sect:hierarchical}, the Hierarchical Linguistic Feature Extractor uses CNN to extract up to 4-gram features, including the unigram features. This is necessary as we assume the 1-size kernel CNN used to extract unigram is used to project the pre-trained word embedding features into the latent space of PAT where it finds easier to fuse information with features from images. In Table \ref{tab:ablation_cnn}, we proved our assumption where PAT which uses an additional 1-size kernel CNN has a better result than one using unigram features extracted from the pre-trained word embedding.
\section{Conclusion and future works}
In this paper, we present the PAT, which achieved the best performance on the benchmark ViVQA dataset. Our ablation study showed that the proposed Hierarchical Linguistic Feature Extractor performed better than LSTM when extracting features from questions.

In future works, we continue to investigate the impact of using Large Language Models (LLMs) on the results of VQA methods, as well as find the most effective multimodal structure that yields the best accuracy on the ViVQA dataset. In addition, our proposed method can be evaluated on two benchmarks datasets: EVJVQA \cite{nguyen2023vlsp} and OpenViVQA \cite{nguyen2023openvivqa}.

\bibliography{main}

\bibliographystyle{abbrv}
%\bibliographystyle{cas-model2-names}

%\bibliography{bibliography}

\end{document}